\documentclass{llncs}
\usepackage{makeidx}  
\usepackage{siunitx}
\usepackage{url}
\usepackage{booktabs}
\usepackage{graphicx}
\usepackage{subcaption}

\newcommand{\bftab}{\fontseries{b}\selectfont}

\usepackage{color}

\begin{document}
\frontmatter          
\pagestyle{headings}  

\mainmatter              
%
\title{An Exploration of 2D and 3D Deep Learning Techniques for Cardiac MR Image Segmentation}
%
%
\author{Christian F. Baumgartner\thanks{both authors contributed equally}\inst{1}, Lisa M. Koch\protect\footnotemark[1]\inst{2}, Marc Pollefeys\inst{2}, Ender Konukoglu\inst{1}}
%
%
\institute{Computer Vision Lab, ETH Zurich \and Computer Vision and Geometry Group, ETH Zurich }

\maketitle              

\begin{abstract}
Accurate segmentation of the heart is an important step towards evaluating cardiac function. In this paper, we present a fully automated framework for segmentation of the left (LV) and right (RV) ventricular cavities and the myocardium (Myo) on short-axis cardiac MR images. 
We investigate various 2D and 3D convolutional neural network architectures for this task.
%
Experiments were performed on the ACDC 2017 challenge training dataset comprising cardiac MR images of 100 patients, where manual reference segmentations were made available for end-diastolic (ED) and end-systolic (ES) frames.
We find that processing the images in a slice-by-slice fashion using 2D networks is beneficial due to a relatively large slice thickness. However, the exact network architecture only plays a minor role. 
We report mean Dice coefficients of $0.950$ (LV), $0.893$ (RV), and $0.899$ (Myo), respectively with an average evaluation time of 1.1 seconds per volume on a modern GPU.


\end{abstract}
\section{Introduction}

Cardiovascular diseases are a major public health concern and currently the leading cause of death in Europe~\cite{Nichols2014}. Automated segmentation of cardiac structures from medical images is an important step towards analysing normal and pathological cardiac function on a large scale, and ultimately towards developing diagnosis and treatment methods.

Until recently, the field of anatomical segmentation was dominated by atlas-based techniques~(e.g. \cite{Bai2015a}), which have the advantage of providing strong spatial priors and yielding robust results with relatively little training data. With more data becoming available and recent advances in machine learning and parallel computing infrastructure, segmentation techniques based on deep convolutional neural networks (CNN) are emerging as the new state-of-the-art~\cite{Ronneberger2015a,Litjens2017}.


This paper is dedicated to the segmentation of cardiac structures on short-axis MR images and is accompanied by a submission to the automated cardiac diagnosis challenge (ACDC) 2017. Short-axis MR images consist of a stack of 2D MR images acquired over multiple cardiac cycles which are often not perfectly aligned and typically have a low through-plane resolution of $5-10$\,mm. 

In this paper, we investigate the suitability of state-of-the-art 2D and 3D CNNs for the segmentation of three cardiac structures. A specific focus is to answer the question if 3D context is beneficial for this task in light of the low through-plane resolution. Furthermore, we explore different network architectures and employ a variety of techniques which are known to enhance training and inference performance in deep neural network such as batch normalisation~\cite{Ioffe}, and different loss functions~\cite{Milletari2016}.
The proposed framework was evaluated on the training set for the ACDC 2017 segmentation challenge. Accurate segmentation results were obtained with a fast inference time of 1.1\,s per 3D image.


\section{Method}


In the following, we will outline the individual steps focusing on the pre-processing, network architectures, optimisation and post-processing of the data. 

\subsection{Pre-Processing}\label{sec:preprocessing}

Since the data were recorded at varying resolutions, we resampled all images and segmentations to a common resolution. For the networks operating in 2D, the images were resampled to an in-plane resolution of $1.37\times1.37$\,\si{\milli\meter}. We did not perform any resampling in the through-plane direction to avoid any losses in accuracy in the up- and downsampling steps. Part of the data had a relatively low through-plane resolution of 10\,\si{\milli\meter} and we found that losses incurred by resampling artefacts can be significant. For the 3D network we chose a resolution of $2.5 \times 2.5 \times 5$\,\si{\milli\meter}. Higher resolutions were not possible due to GPU memory restrictions. We then placed all the resampled images centrally into images of constant size, padding with zeros where necessary. The exact image size depended on the network architecture and will be discussed below. 
%
Lastly, each image was intensity-normalised to zero mean and unit variance.

\subsection{Network Architectures}\label{sec:networks}

We investigated four different network architectures. The fully convolutional segmentation network (FCN) proposed by~\cite{Long} is a 2D segmentation network widely used for natural images. In this architecture deep, and thus coarse, feature maps are upsampled to the original image resolution by using transposed convolutions. In order to fuse the semantic information available in the deeper layers with the spatial information available in the shallower stages, the authors proposed to use skip connections. In the present work, we used the best performing incarnation which is based on the VGG-16 architecture and uses three skip connections (FCN-8)~\cite{Long}. We used an image size of $224\times224$ pixels for this architecture. 

Another popular segmentation architecture is the 2D U-Net initially proposed for the segmentation of neuronal structures in electron microscopy stacks and cell tracking in light microscopy images~\cite{Ronneberger2015a}. Inspired by~\cite{Long} the authors employ an architecture with symmetric up- and downsampling paths and skip connections within each resolution stage. Since this architecture does not employ padded convolutions, a larger image size of $396\times396$ pixels was necessary, which led to segmentation masks of size $212\times212$ pixels.

Inspired by the fact that the FCN-8 produces competitive results despite having a simple upsampling path with few channels, we speculated that the full complexity of the U-Net upsampling path may not be necessary for our problem. Therefore, we additionally investigated a modified 2D U-Net with number of feature maps in the transpose convolutions of the upsampling path set to the number of classes. Intuitively, each class should have at least one channel. 

\c{C}i\c{c}ek et al. recently extended the U-Net architecture to 3D~\cite{Cicek2016} by following the same symmetric design principle. However, for data with few slices in one orientation, the repeated pooling and convolving may be too aggressive. We found that using the 3D U-Net for our data all spatial information in the through-plane direction was lost before the third max pooling step.  
%
%
We thus also investigated a slightly modified version of the 3D U-Net in which we performed only one max-pooling (and upsampling) step in the through-plane direction. This had two advantages: 1) The spatial information in the through-plane was retained and thus available in the deeper layers, 2) it allowed us to work with a slightly higher image resolution because less padding in the through-plane direction (and thus less GPU memory) was required. In preliminary experiments we found that the modified 3D U-Net led to improvements of around 0.02 of the average Dice score over the standard 3D U-Net. In the interest of brevity we only included the modified version in the final results of this paper. Here, we used an input image size of $204\times204\times60$, which led to output masks of size $116\times116\times28$. 

We used batch normalisation~\cite{Ioffe} on the outputs of every convolutional and transposed convolutional layer for all architectures. We found that this not only led to faster convergence, as reported in~\cite{Cicek2016}, but also consistently yielded better results and allowed the training of some networks to converge that did not converge otherwise. 

\subsection{Optimisation}\label{sec:optimisation}

We trained the networks introduced above (i.e. FCN-8, 2D U-Net, 2D U-Net (mod.) and 3D U-Net (mod.)) from scratch with the weights of the convolutional layers initialised as described in~\cite{He2015}.

%
We investigated three different cost functions. First, we used the standard pixel-wise cross entropy. 
%
To account for the class imbalance between the background and the foreground classes, we also investigated a weighted cross entropy loss. We used a weight of $0.1$ for the background class, and $0.3$ for the foreground classes in all experiments in this paper, which corresponds approximately to the inverse prevalence of each label in the dataset. 
%
Lastly, we investigated optimising the Dice coefficient directly. In order to get more stable gradients we calculated the Dice loss on the softmax output as follows:
$$ \mathcal{L}_{dice} = 1 - \frac{\sum_{k=2}^K \sum_{n=1}^N t_{nk}y_{nk}}{\sum_{k=2}^K \sum_{n=1}^N t_{nk} + y_{nk}}, $$
where $K$ is the number of classes, $N$ the number of pixels/voxels, $y$ is the softmax output, $t$ is a one-hot vector encoding the true label per location. 

To minimise the respective cost functions we used the ADAM optimiser~\cite{Kingma2015} with a learning rate of 0.01, $\beta_1=0.9$ and $\beta_2=0.999$. The best results were obtained without using any weight regularisation. The training of each of the models took approximately 24 hours on a Nvidia Titan Xp GPU. 

\subsection{Post-Processing}

Since training and inference were performed in a different resolution, the predictions had to be resampled to each subject's initial resolution. To avoid resampling artefacts, this step was carried out on the softmax (i.e. continuous) network outputs for each label using linear interpolation. The final discrete segmentation was then obtained in the final resolution by choosing the label with the highest score at each voxel. Interpolation on the softmax output, rather than the output masks, led to consistent improvements of around 0.005 in the average Dice score. 

We occasionally observed spurious predictions of structures in implausible locations. To compensate for this, we applied simple post-processing to the segmentation results by keeping only the largest connected component for every structure. Since the segmentations are already quite accurate without post-processing this only lead to an average Dice increase of approximately 0.0003, however, it reduced the Hausdorff distance considerably, which by definition is very sensitive to outliers.
Other post-processing techniques such as the commonly used spatial regularisation method based on fully connected conditional random fields~\cite{Krahenbuhl2012} did not yield improvements in our experiments. 
%


\section{Experiments and Results}

\subsection{Data}

The experiments in this paper were performed on cardiac cine-MRI training data of the ACDC challenge\footnote{\url{https://www.creatis.insa-lyon.fr/Challenge/acdc} (last accessed 26 July 2017)}. The publicly available training dataset consists of 100 patient scans each including a short-axis cine-MRI acquired on 1.5T and 3T systems with resolutions ranging from $0.70\times0.70$\,\si{\milli\meter} to $1.92\times1.92$\,\si{\milli\meter} in-plane and $5$\,\si{\milli\meter} to 10\,\si{\milli\meter} through-plane. Furthermore, segmentation masks for the myocardium (Myo), the left ventricle (LV) and the right ventricle (RV) are available for the end-diastolic (ED) and end-systolic (ES) phases of each patient. The dataset includes, in equal numbers, patients diagnosed with previous myocardial infarction, dilated cardiomyopathy, hypertrophic cardiomyopathy, abnormal right ventricles, as well as normal controls. 
%
We did not employ any external data for training or pre-training of the networks.

The dataset was divided into a training and validation set comprising 80 and 20 subjects, respectively, with a stratified split w.r.t. patient diagnosis. All images were pre-processed as described in Sec.~\ref{sec:preprocessing}.

\subsection{Evaluation Measures}

We evaluated the segmentation accuracy achieved with the different network architectures and optimisation techniques using three measures: the Dice coefficient, the Hausdorff distance and the average symmetric surface distance (ASSD). 
Furthermore, for the best performing experiment configuration, the correlations to commonly measured clinical variables were calculated.


\subsection{Experiment 1: Comparison of Loss Functions}

In the first experiment we focused on the modified 2D U-Net architecture for which we obtained good initial results, and compared the performance using the different cost functions introduced in Sec.~\ref{sec:optimisation}. In Table~\ref{tab:optimisation_results} we report the Dice score and ASSD averaged over both cardiac phases. It can be seen that using cross entropy led to better results than optimising the Dice directly.
Weighted and unweighted cross entropy performed similarly, with the weighted loss function leading to marginally better results. We conclude that for the task at hand, the class imbalance does not seem to be an issue. Nevertheless, for the comparison of the network architectures in the next section we continued using the unweighted cross entropy as a loss function due to the slightly better results.

\begin{table}[t]
   \caption{Segmentation accuracy obtained by optimising the modified 2D U-Net using different cost functions.}
 \centering
 \resizebox{\columnwidth}{!}{%
\begin{tabular}{lcccccccc}
           
  \toprule
           & Dice (LV) & ASSD (LV) & ~ & Dice (RV) & ASSD (RV) & ~ & Dice (Myo) & ASSD (Myo) \\
  \midrule
  Crossentropy  & \bftab 0.950\,(0.029)  & \bftab 0.43\,(0.41)  &  & 0.891\,(0.084)  & 1.06\,(1.04)  &  & 0.888\,(0.031)  & 0.52\,(0.22)  \\ 
  W. Crossentropy  & \bftab 0.950\,(0.036)  & 0.52\,(0.75)  &  & \bftab 0.893\,(0.083)  & \bftab 1.04\,(1.06)  &  & \bftab 0.899\,(0.032)  & \bftab 0.51\,(0.35)  \\
  Dice Loss  & 0.944\,(0.051)  & 0.56\,(0.77)  &  & 0.843\,(0.137)  & 2.13\,(2.03)  &  & 0.891\,(0.029)  & 0.55\,(0.24)  \\
  \bottomrule
\end{tabular}%
}
   \label{tab:optimisation_results}
\end{table}

\subsection{Experiment 2: Comparison of Network Architectures}

This experiment focuses on the comparison of the different 2D and 3D network architectures described in Sec.~\ref{sec:networks}. The results are shown in Table~\ref{tab:optimisation_results}. 
It can be seen that the 2D U-Net (both the original and modified version) outperformed FCN-8 and the (modified) 3D U-Net. While both versions of the 2D U-Net perform similarly, the modified version leads to slightly better results. 

\begin{figure}[t]
\centering
\includegraphics[width=\textwidth]{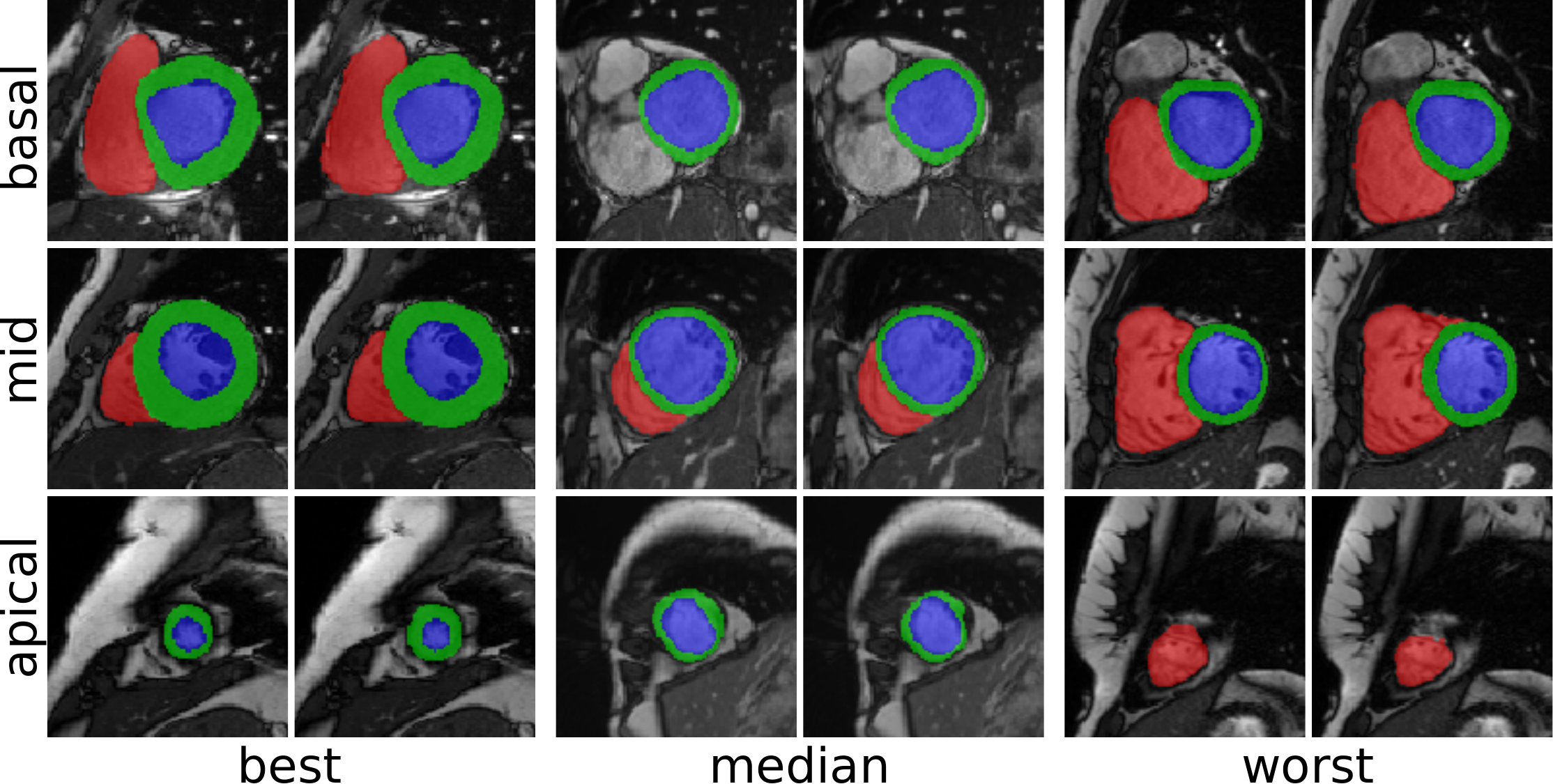} 
\caption{Example segmentations at ED obtained using the 2D U-Net (mod.) for subjects with the highest, median, and lowest Dice coefficients on the Myocardium (left to right). Ground truth (left) and predicted segmentation (right) are shown for a basal, mid-ventricular and apical slice (top to bottom).}
\label{fig:qualresults}
\end{figure}

Clinical measures for the best performing method (the modified 2D U-Net) are shown in Table~\ref{tab:clinical_measures}. A detailed description of the measures is provided by ACDC$^{3}$. Figure~\ref{fig:qualresults} shows example segmentation results at three slice positions using the above method.
%
%
Inference on a single volume took approximately 1.1\,\si{\second} for the 2D networks and 2.2\,\si{\second} for the 3D networks using a Nvidia Titan Xp GPU. 

\begin{table}[t]
   \caption{Segmentation accuracy measures for different network architectures. Each table entry depicts the mean (std) value accuracy measure obtained for a specific structure and cardiac phase.}
 \centering
 \resizebox{\columnwidth}{!}{%
\begin{tabular}{lccccccc}

  \toprule
                  & \multicolumn{3}{c}{Left Ventricle (ED)}     & ~ & \multicolumn{3}{c}{Left Ventricle (ES)} \\
\cmidrule(lr){2-4}\cmidrule(lr){6-8} 
                  & Dice          & ASSD          & HD            & & Dice        & ASSD          & HD  \\
  \midrule
  FCN-8           & 0.960\,(0.018)  & 0.41\,(0.49)  & 5.77\,(3.05)  &  & 0.926\,(0.061)  & 0.64\,(0.80)  & 7.31\,(3.39)  \\ 
  2D U-Net        & 0.965\,(0.014)  & \bftab 0.36\,(0.38)  & \bftab 5.63\,(2.79)  &  & \bftab 0.937\,(0.051)  & \bftab 0.54\,(0.64)  & \bftab 6.85\,(3.52)  \\ 
  2D U-Net (mod.) & \bftab 0.966\,(0.017)  & 0.37\,(0.48)  & 5.71\,(4.22)  &  & 0.935\,(0.042)  & 0.67\,(0.92)  & 8.23\,(8.29)  \\ 
  3D U-Net (mod.) & 0.939\,(0.022)  & 0.63\,(0.50)  & 8.69\,(4.25)  &  & 0.905\,(0.039)  & 0.70\,(0.38)  & 9.13\,(4.10)  \\ 

  \midrule
                  & \multicolumn{3}{c}{Right Ventricle (ED)}     & ~ & \multicolumn{3}{c}{Right Ventricle (ES)} \\
\cmidrule(lr){2-4}\cmidrule(lr){6-8} 
                  & Dice          & ASSD          & HD            & & Dice        & ASSD          & HD  \\
  \midrule
  FCN-8           & 0.932\,(0.025)  & \bftab 0.57\,(0.45)  & 12.24\,(5.51)  &  & 0.835\,(0.100)  & 1.63\,(1.07)  & 13.89\,(4.24)  \\
  2D U-Net        & \bftab 0.936\,(0.028)  & 0.65\,(0.48)  & 12.43\,(6.13)  &  & 0.838\,(0.085)  & 1.72\,(1.22)  & 14.52\,(5.28)  \\
  2D U-Net (mod.) & 0.934\,(0.039)  & 0.66\,(0.74)  & \bftab 12.17\,(6.02)  &  & \bftab 0.852\,(0.095)  & \bftab 1.42\,(1.19)  & \bftab 13.46\,(6.24)  \\  
  3D U-Net (mod.) & 0.888\,(0.069)  & 1.17\,(1.21)  & 14.91\,(5.02)  &  & 0.781\,(0.101)  & 2.26\,(1.40)  & 16.24\,(5.39)  \\ 

  \midrule
                  & \multicolumn{3}{c}{Myocardium (ED)}     & ~ & \multicolumn{3}{c}{Myocardium (ES)} \\
\cmidrule(lr){2-4}\cmidrule(lr){6-8} 
                  & Dice          & ASSD          & HD            & & Dice        & ASSD          & HD  \\
  \midrule
  FCN-8           & 0.869\,(0.029)  & 0.55\,(0.23)  & 9.16\,(6.74)  &  & 0.890\,(0.027)  & 0.62\,(0.24)  & 9.69\,(5.28)  \\ 
  2D U-Net        & 0.885\,(0.027)  & 0.52\,(0.29)  & 9.01\,(7.66)  &  & 0.904\,(0.029)  & \bftab 0.55\,(0.28)  & 10.06\,(5.79)  \\
  2D U-Net (mod.) & \bftab 0.892\,(0.027)  & \bftab 0.45\,(0.22)  & \bftab 8.65\,(6.02)  &  & \bftab 0.906\,(0.034)  & 0.56\,(0.44)  & \bftab 9.66\,(6.21)  \\ 
  3D U-Net (mod.) & 0.802\,(0.053)  & 0.91\,(0.34)  & 11.87\,(6.25)  &  & 0.839\,(0.066)  & 0.90\,(0.42)  & 10.95\,(3.47)  \\ 

  \bottomrule 

\end{tabular}%
}
   \label{tab:optimisation_results}
\end{table}

\begin{table}[t]
   \caption{Clinical measurements: correlation, bias and limits of agreement (LoA) for the LV and RV ejection fraction (EF) and all structure volumes.}
 \centering
 \resizebox{\columnwidth}{!}{%
\begin{tabular}{lccccccc}          
  \toprule
           & \multicolumn{3}{c}{Correlation}     & ~ & \multicolumn{3}{c}{Bias [LoA]} \\
\cmidrule(lr){2-4}\cmidrule(lr){6-8}
           & EF & Vol (ED) & Vol (ES) & ~ & EF & Vol (ED) & Vol (ES) \\
  \midrule
  LV  & 0.972 &	0.998 &	0.994 & & 	$-0.45 [-9.68 ; 8.78]$	& $1.28	[-7.27 ; 9.83]$	&	$2.55 [-14.86 ; 19.96]$       \\ 
  RV  & 0.868 &	0.961 &	0.965 & &  $6.25 [-13.08 ; 25.58]$ & $-0.45 [-28.89 ; 27.99]$ &	$-8.08 [-33.87 ; 17.71]$    \\ 
  Myo  &  - & 0.995 & 0.988    & & - & $-5.24 [-15.27 ; 4.79]$	& $-0.71 [-17.89 ; 16.47]$   		       \\ 
  \bottomrule
\end{tabular}%
}
   \label{tab:clinical_measures}
\end{table}

\subsection{Discussion and Conclusion}

In this work we evaluated the suitability of state-of-the-art neural network architectures for the task of fully automatic cardiac segmentation. We also investigated modified versions of those networks which yielded marginal improvements in performance. In particular, we found that using fewer feature maps in the upsampling path of the 2D U-Net yielded minor but consistent improvements. We speculate that for this problem the full complexity of the upsampling path is not necessary. Furthermore, the ``bottlenecks'' may force the downsampling layers to learn more semantically meaningful features. Lastly, having fewer parameters may also make the problem easier to optimise. Further investigation into the significance of the upsampling path complexity will be necessary. 

Overall we found that the exact architecture played a minor role in the accuracy of the system. However, the use of batch normalisation as well as the choice of the cost function had a big impact on the performance. Moreover, we found that resampling of the predictions to the original image resolution was a significant source of errors. This could be reduced by resampling the softmax output with linear interpolation, rather than the predicted masks. 

One goal of this paper was to investigate if 3D context is helpful for the segmentation of short-axis MR images. Our experiments revealed that all 2D approaches consistently outperformed the (modified) 3D U-Net. There are at least three possible reasons for this: (1) when using 3D data, the amount of training images is drastically reduced which complicates training. (2) Since the through-plane resolution is low (and the cardiac structures typically appear in the top and bottom slices already), border effects from 3D convolutions may compromise the information available at intermediate representations. (3) GPU memory restrictions required a substantial downsampling of the data for training and prediction, potentially leading to a loss of information.

The segmentation scores reported in this work compare favourably to the related literature. However, it should be noted that a direct comparison is complicated by the fact that different datasets were used in the different works. For the LV cavity two recent deep learning methods~\cite{Avendi2016,Oktay2017} report Dice scores of around 0.94, while the modified 2D U-Net discussed here achieved a slightly higher value of 0.95. For automated segmentation of the RV cavity, \cite{Bai2013,Oktay2017a} report similar results to ours. Segmentation of the myocardium is a more challenging task than the LV and RV cavities, which is reflected by lower Dice scores of around 0.81 reported in recent literature~\cite{Bai2015a,Oktay2017}. We achieved substantially higher results using all 2D architectures. In particular, the modified 2D U-Net architecture produced a Dice score of 0.899 for this structure. While these results are encouraging, further analysis on common datasets is necessary. Specifically, we observed that the field of view in many images of the ACDC challenge dataset does not include the apex and basal region of the heart, which are particularly challenging to segment. 

The code and pretrained models for all examined network architectures are publicly available at \url{https://github.com/baumgach/acdc_segmenter}.
\bibliographystyle{splncs03}
\bibliography{references_stripped}

\begin{thebibliography}{10}
\providecommand{\url}[1]{\texttt{#1}}
\providecommand{\urlprefix}{URL }

\bibitem{Avendi2016}
Avendi, R.M.R., Kheradvar, A., Jafarkhani, H.: {A combined deep-learning and
  deformable-model approach to fully automatic segmentation of the left
  ventricle in cardiac MRI}. Med Image Anal  30,  108--119 (2016)

\bibitem{Bai2015a}
Bai, W., Shi, W., Ledig, C., Rueckert, D.: {Multi-atlas segmentation with
  augmented features for cardiac MR images.} Med Image Anal  19(1),  98--109
  (2015)

\bibitem{Bai2013}
Bai, W., Shi, W., O'Regan, D.P., Tong, T., Wang, H., Jamil-Copley, S., Peters,
  N.S., Rueckert, D.: {A probabilistic patch-based label fusion model for
  multi-atlas segmentation with registration refinement: application to cardiac
  MR images.} IEEE Transactions on Medical Imaging  32(7),  1302--15 (2013)

\bibitem{Cicek2016}
{\c{C}}i{\c{c}}ek, {\"{O}}., Abdulkadir, A., Lienkamp, S.S., Brox, T.,
  Ronneberger, O.: {3D U-Net: Learning Dense Volumetric Segmentation from
  Sparse Annotation}. In: MICCAI. pp. 424--432 (2016)

\bibitem{He2015}
He, K., Zhang, X., Ren, S., Sun, J.: {Delving Deep into Rectifiers: Surpassing
  Human-Level Performance on ImageNet Classification}. In: ICCV. pp. 1026--34
  (2015)

\bibitem{Ioffe}
Ioffe, S., Szegedy, C.: {Batch Normalization: Accelerating Deep Network
  Training by Reducing Internal Covariate Shift}. In: ICML. pp. 448--456 (2015)

\bibitem{Kingma2015}
Kingma, D.P., Ba, J.L.: {ADAM: A Method for Stochastic Optimization}. In: ICLR
  (2015)

\bibitem{Krahenbuhl2012}
Kr{\"{a}}henb{\"{u}}hl, P., Koltun, V.: {Efficient Inference in Fully Connected
  CRFs with Gaussian Edge Potentials}. In: NIPS. pp. 109--117 (2011)

\bibitem{Litjens2017}
Litjens, G., Kooi, T., Bejnordi, B.E., Setio, A.A.A., Ciompi, F., Ghafoorian,
  M., van~der Laak, J.A.W.M., van Ginneken, B., S{\'{a}}nchez, C.I.: {A Survey
  on Deep Learning in Medical Image Analysis}. arXiv:1702.05747  (2017)

\bibitem{Long}
Long, J., Shelhamer, E., Darrell, T.: {Fully Convolutional Networks for
  Semantic Segmentation}. In: CVPR. pp. 343 --3440 (2015)

\bibitem{Milletari2016}
Milletari, F., Navab, N., Ahmadi, S.A.: {V-Net: Fully Convolutional Neural
  Networks for Volumetric Medical Image Segmentation}. In: 3D Vision. pp. 565
  -- 571 (2016)

\bibitem{Nichols2014}
Nichols, M., Townsend, N., Scarborough, P., Rayner, M.: {Cardiovascular disease
  in Europe 2014 : epidemiological update}. European heart journal  (2014)

\bibitem{Oktay2017a}
Oktay, O., Bai, W., Guerrero, R., Rajchl, M., de~Marvao, A., O'Regan, D.P.,
  Cook, S.A., Heinrich, M.P., Glocker, B., Rueckert, D.: {Stratified Decision
  Forests for Accurate Anatomical Landmark Localization in Cardiac Images}.
  IEEE Trans Med Imag  36(1),  332--342 (2017)

\bibitem{Oktay2017}
Oktay, O., Ferrante, E., Kamnitsas, K., Heinrich, M., Bai, W., Caballero, J.,
  Guerrero, R., Cook, S., de~Marvao, A., Dawes, T., O'Regan, D., Kainz, B.,
  Glocker, B., Rueckert, D.: {Anatomically Constrained Neural Networks (ACNN):
  Application to Cardiac Image Enhancement and Segmentation}. arXiv:1705.08302
  (2017)

\bibitem{Ronneberger2015a}
Ronneberger, O., Fischer, P., Brox, T.: {U-Net: Convolutional Networks for
  Biomedical Image Segmentation}. In: MICCAI. pp. 234--241 (2015)

\end{thebibliography}

\end{document}